\documentclass[letterpaper, 10 pt, conference]{ieeeconf}  

\usepackage[T1]{fontenc}
\usepackage{wrapfig}
\usepackage{booktabs}
\usepackage{amsmath}
\usepackage{amsmath} 
\usepackage{amssymb}  
\usepackage[]{changes}
\definechangesauthor[name={Todor}, color=red]{tsv}
\definechangesauthor[name={Johannes}, color=cyan]{jsk}
\presetkeys%
    {todonotes}%
    {inline,backgroundcolor=yellow}{}
\usepackage{chngcntr}
\usepackage{dcolumn}
\newcolumntype{d}[1]{D{.}{\cdot}{#1} }
\usepackage{siunitx}
\usepackage{tikz}
\usepackage{adjustbox}

\newcommand{\citep}[1]{\cite{#1}}
\newcommand{\citet}[1]{\cite{#1}}

\IEEEoverridecommandlockouts                              

\title{\LARGE \bf
Ensemble of Sparse Gaussian Process Experts for\\
Implicit Surface Mapping with Streaming Data
}

\author{Johannes A. Stork and Todor Stoyanov
\thanks{The authors are with the Center for Applied Autonomous Sensor Systems (AASS) and Autonomous Mobile Manipulation Lab, Örebro University, Sweden
{\tt\small [johannes.stork | todor.stoyanov]@oru.se}}%
}

\begin{document}

\maketitle
\thispagestyle{empty}
\pagestyle{empty}

\begin{abstract}
Creating maps is an essential task in robotics and provides the basis for effective planning and navigation. 
In this paper, we learn a compact and continuous implicit surface map of an environment from a stream of range data with known poses.
For this, we create and incrementally adjust an ensemble of approximate Gaussian process (GP) experts which are each responsible for a different part of the map. 
Instead of inserting all arriving data into the GP models, we greedily trade-off between model complexity and prediction error. Our algorithm therefore uses less resources on areas with few geometric features and more where the environment is rich in variety. 
We evaluate our approach on synthetic and real-world data sets and analyze sensitivity to parameters and measurement noise. The results show that we can learn compact and accurate implicit surface models under different conditions, with a performance comparable to or better than that of exact GP regression with subsampled data. 
\end{abstract}


\section{\uppercase{Introduction}}
\label{sec:introduction}

The standard robotics approach to mapping with known poses is estimating a discrete model of occupancy probabilities~\citep{Moravec_and_Elfes_1985}. While being straight-forward, it offers poor generalization because cells are considered independently. In contrast, \textit{continuous} models based on implicit surface representation can interpolate between sparse observations and make use of prior information \cite{williams2007gaussian}. However, current state-of-the-art approaches do not scale well in terms of data or the size of the covered space \cite{kim2014recursive, wang2016fast, jadidi2018gaussian}. In fact, fundamental questions about the underlying representations and algorithms for learning still need to be researched more before continuous mapping can become relevant in practice. A key open challenge for applying continuous mapping to robotics scenarios is that measurements arrive in sequence, making it necessary to incrementally construct and update the map efficiently.

\begin{figure}
\centering

\hspace*{\fill}
\includegraphics[height=0.18\linewidth]{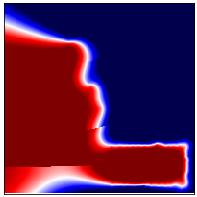}
\hfill
\includegraphics[height=0.18\linewidth]{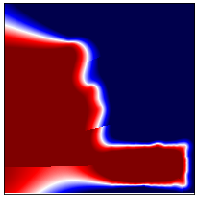}
\hfill
\includegraphics[height=0.18\linewidth]{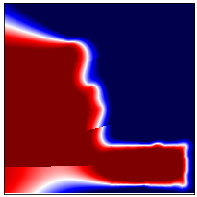}
\hfill
\includegraphics[height=0.18\linewidth]{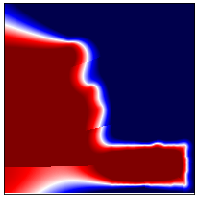}
\hfill
\includegraphics[height=0.18\linewidth]{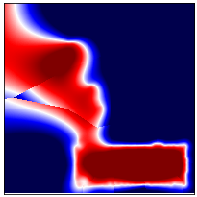}
\hspace*{\fill}

\hspace*{\fill}
\includegraphics[height=0.18\linewidth]{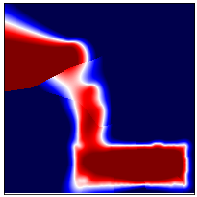}
\hfill
\includegraphics[height=0.18\linewidth]{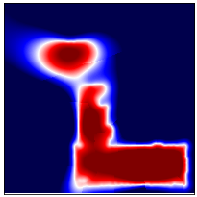}
\hfill
\includegraphics[height=0.18\linewidth]{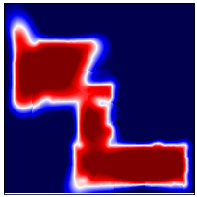}
\hfill
\includegraphics[height=0.18\linewidth]{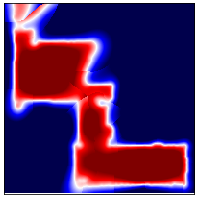}
\hfill
\includegraphics[height=0.18\linewidth]{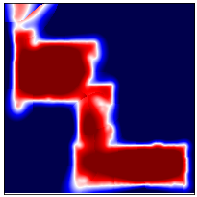}
\hspace*{\fill}

\hspace*{\fill}
\includegraphics[height=0.31\linewidth]{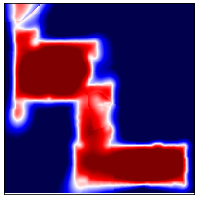}
\hfill
\includegraphics[height=0.31\linewidth]{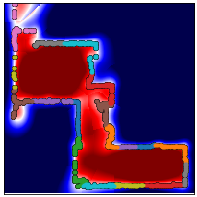}
\hfill
\includegraphics[height=0.31\linewidth]{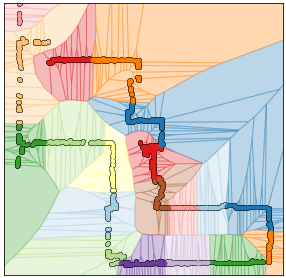}
\hspace*{\fill}

\caption{%
Learning of the singled distance function: Map prediction after the first 10 learning steps from real-world data set \emph{Basement-real} (top two rows). Prediction of the final signed distance function (bottom left), locations of the primary pseudo-inputs of local experts in different colors (bottom middle), and the areas of responsibility of experts (bottom right).}
\label{fig:title}
\vspace{-0.8cm}
\end{figure}

In this article, we propose a method for online continuous map learning in a streaming data setting, based on an ensemble of sparse pseudo-input Gaussian processes (GP). For this, we build on the Gaussian process implicit surface (GPIS)~\citep{williams2007gaussian} representation, which encodes surface locations as zero crossings of a continuous signed distance function. 
We address the computational infeasibility of maintaining an exact GP regression of large amounts of data by: (1) data-driven partitioning of the map into disjoint regions of responsibility; (2) forming an ensemble of sparse GP local experts; and (3) devising a method for ensuring the continuity between expert predictions. The proposed method results in a compact implicit surface representation that uses more resources (i.e. pseudo-inputs) in areas of high geometric texture while maintaining sparsity in areas with uniform structure. 
When adjusting our ensemble with new data, our sparse local GP models are updated, dynamically extended when necessary, contracted when possible, and divided-up in a greedy trade-off between computational feasibility and prediction error. An example is depicted in Fig.~\ref{fig:title}.
The main contributions of this work are thus: (1) a method for online fitting of continuous surface models from streaming data; (2) a greedy method for adding and removing pseudo-inputs in sparse GP models; (3) a greedy method for subdividing sparse GP models that maintain continuous predictions.

\section{\uppercase{Related Work}}
\label{sec:relatedwork}

Discrete grid-based estimation methods, such as Occupancy Grid Maps~\citep{Moravec_and_Elfes_1985}, Normal Distribution Transforms~\citep{Biber_and_Strasser_2003}, or discrete Signed Distance Fields (SDFs)~\citep{curless1996volumetric} are often use in robotics. While these methods produce good results and are straight-forward to implement, they offer poor generalization and some suffer from discretization artifacts. 
Continuous map models encode surfaces as a parameterized function which can give more compact models and better generalization performance. The most advanced approaches of this type use multivariate Gaussian mixture models (GMM)\cite{srivastava2016approximate, evangelidis2014generative, eckart2013rem}, nonlinear logistic discriminative models called Hilbert maps (HM) \cite{ramos2016hilbert, doherty2016probabilistic, guizilini2016large}, and Gaussian processes (GP) \cite{o2012gaussian, kim2015gpmap, jadidi2014exploration}. 
A common challenge of these methods is online processing of data: GMMs typically require iterative expectation maximization for model initialization and update~\citep{srivastava2016approximate}; HMs can be constructed incrementally~\citep{senanayake2017bayesian}, but cannot be updated online in a straight-forward manner; and GP-based models become computationally infeasible as data grows large~\citep{kim2014recursive, jadidi2014exploration, jadidi2018gaussian}. 

While GP-based models have been applied to occupancy mapping~\citep{kim2013continuous, jadidi2014exploration, o2009contextual}, we learn an implicit surface model~\citep{williams2007gaussian, kim2014recursive, kim2015gpmap, kim2015hierarchical} with a signed distance function (see Sec.~\ref{sec:gpis} for details). 
In most cases, the prohibitive cost of GP regression for large data is addressed by partitioning the data and fitting separate local GPs~\citep{kim2013continuous, kim2015hierarchical, kim2015gpmap, kim2014recursive, o2009contextual}. Continuous predictions are then provided by overlapping data~\citep{kim2013continuous}, Bayesian committee machines (BCM)~\citep{kim2014recursive}, or hierarchical modeling~\citep{kim2015hierarchical}. We focus on the problem of online learning from streaming data where only a small amount of data is available at any given time and use sparse pseudo-input Gaussian processes~\citep{quinonero2005unifying}.  
Only a handful of the GP-based mapping methods are capable of online learning as required for streaming data: Most of them use the BCM to recursively merge posterior predictions for map locations on a predefined grid~\citep{kim2014recursive, jadidi2018gaussian}. While this results in computational gains with respect to GP regression, it requires the same amount of resources in all regions of the map, empty or not, and can lead to model divergence over time.

More similar to our approach, \citet{lee2019online} propose a method to update a GPIS model by manipulating the already present measurement data in the model. For this, they present a recursive Bayesian update for single measurements. In their case, model size is controlled by a spatial data structure which is used to subdivide models with too many points. Different to our method, arriving data is always inserted to the model (up to a minimal distance threshold between measurements). While this bounds the size of local GP models, it does not necessarily make efficient use of resources as measurements in the model are distributed uniformly around surface locations and do not adapt to the complexity of surface geometry. Continuous predictions are achieved by sharing data between neighboring GP models and posterior predictions are combined in a weighted sum. As a consequence, updating a single measurement requires re-computation of all affected models. 
In our model, pseudo-inputs are maintained at shared locations to achieve overlap for continuous prediction, but they can be updated independently for each model. While their approach partitions space by a predefined spatial pattern, we rely on data-driven partitioning. 

\section{\uppercase{Implicit Surface Estimation as\\ Gaussian Process Regression}}
\label{sec:gpis}

Our goal is to estimate surface locations in an environment in form of a signed distance field $d \colon \mathbb{R}^2 \to \mathbb{R}$ (SDF) which implicitly represents the surface as the zero set $d^{-1}(0) = \{ \mathbf{x} \in \mathbb{R}^2 | d(\mathbf{x}) = 0\}$. At other locations, the SDF is positive when outside of the surface and negative when behind the surface (e.g.\ inside an object). For regression of the SDF, we obtain measurements $(\mathbf{x}, y)$ consisting of locations $\mathbf{x} \in \mathbb{R}^2$ and signed distance values $y \in \mathbb{R}$. Range sensor data with known sensor pose provides us with noisy surface measurements at a set of locations $X_\mathrm{surf}$ with distance value $0$. Given the sensor pose, we construct additional auxiliary measurements $X_\mathrm{aux}$ which are close to the surface and have a defined positive distance value, e.g.\ $0.1$ in front of the surface along the sensor rays (akin to a projective SDF~\citep{curless1996volumetric}). We summarize measurements in shorthand notation as the set of measurement locations $X_m = X_\mathrm{surf} \cup X_\mathrm{aux}$ with $[X_m]_i = \mathbf{x}_i$ and the output vector of signed distance values $\mathbf{y}$ with $[\mathbf{y}]_i = y_i$.

For GP regression of the SDF, we assume that the measured data are corrupted with noise, such that $y_i = f(\mathbf{x}_i) + \varepsilon$, with $\varepsilon \sim \mathcal{N}(0, \sigma^2)$. To predict the SDF at trail locations $\mathbf{x}_* \in X_*$, we first select a joint prior distribution over function values $\mathbf{f}_m = f(X_m)$ and $\mathbf{f}_* = f(X_*)$. GPIS uses a prior distribution  derived from thin plate spline regularization \citep{williams2007gaussian} with the kernel $k(\mathbf{x}_1, \mathbf{x}_2) = 2r^2 \log |r| - (1+ 2 \log(R))r^2 + R^2$, where $r = ||\mathbf{x}_1 - \mathbf{x}_2||_2$ and $R$ is the largest distance in the environment. This kernel function minimizes the second order derivative of the  posterior function which is desirable for smooth surfaces estimation. With exact GP regression, the posterior distribution for SDF values is $\mathbf{f}_* \sim \mathcal{N}(\boldsymbol{\mu}_*, \Sigma_{**})$, where $\boldsymbol{\mu}_* = K_{*m} (K_{mm} + \Sigma_{mm})^{-1}\mathbf{y}$ is the mean and $\Sigma_{**} = K_{**} - K_{*m} (K_{mm} + \Sigma_{mm})^{-1} K_{m*}$ is the covariance matrix. Observation noise enters as the diagonal matrix $\Sigma_{mm} = \sigma^2 \mathbf{I}$ and $K_{mm}$, $K_{**}$, and $K_{*m}$ are kernel matrices. 

This approach to surface estimation processes all measurements in one batch which does not scale well to larger amounts of measurements since a matrix of dimensions $|X_m| \times |X_m|$ is inverted. In the next section, we explain an approximation to GP regression that reduces the computational complexity of predicting $\mathbf{f}_*$ and allows for online processing of measurements.

\section{\uppercase{Sparse Pseudo-Input Gaussian Process and Streaming Data}}
\label{sec:FITC-gp}

In the streaming data setting, we assume that a large amount of measurements arrive one by one in a stream of sensor data and that the regression therefore needs to be constructed incrementally. As a consequence, the exact GP regression approach from Sec. \ref{sec:gpis} becomes practically intractable, both, because of the large amount of data and the repeated re-computation of predictions. We address this problem in two ways: First, we approximate the exact GP regression model (explained in this section) and second, we use data-driven spatial partitioning of the input space into multiple, smaller regression models (see Sec. \ref{sec:method}).

The sparse pseudo-input Gaussian process regression \citep{quinonero2005unifying} uses the inducing input (II) assumption and the fully independent training conditional (FITC) assumption to reduce computational complexity and memory requirements. 
Using a set of freely selected pseudo-input (PI) locations $X_u$ and their corresponding function values $\mathbf{f}_u$, the II assumption states that the function values $\mathbf{f}_m$ and $\mathbf{f}_*$ are conditionally independent given $\mathbf{f}_u$, which factorizes the prior distribution as $p(\mathbf{f}_m, \mathbf{f}_*, \mathbf{f}_u) = p(\mathbf{f}_m | \mathbf{f}_u) p(\mathbf{f}_* | \mathbf{f}_u) p(\mathbf{f}_u)$. As a result, the posterior distribution over $\mathbf{f}_*$ is now $\mathcal{N}(\boldsymbol{\mu}_*, \Sigma_{**})$ with 
\begin{gather}
\boldsymbol{\mu}_* = K_{*u} K_{uu}^{-1}\boldsymbol{\mu}_u
\quad \text{and}
\notag
\\
\Sigma_{**} = K_{**} - K_{*u} K_{uu}^{-1} (K_{uu} - \Sigma_{uu}) K_{uu}^{-1} K_{u*}
\label{eq:fitc-prediction}
\end{gather}
as mean and covariance, and the diagonal matrix $\Sigma_{uu} = \sigma^2 \mathbf{I}$ captures the noise. If the number of pseudo-inputs $|X_u|$ is much smaller than the number of measurements $|X_m|$, this results in a much faster prediction for trail points when $\boldsymbol{\mu}_u$ and $\Sigma_{uu}$ are given.

To compute $\boldsymbol{\mu}_u$ and $\Sigma_{uu}$ efficiently, the FITC assumption states that the function values $\mathbf{f}_m$ are independent from each other given $\mathbf{f}_u$, which factorizes the prior distribution as $p(\mathbf{f}_m, \mathbf{f}_*, \mathbf{f}_u)= (\prod_{\mathbf{x} \in X_m} p(f(\mathbf{x}) | \mathbf{f}_u)) p(\mathbf{f}_* | \mathbf{f}_u) p( \mathbf{f}_u)$. As a result, the posterior distribution over $\mathbf{f}_u$ is the Gaussian $\mathcal{N}(\boldsymbol{\mu}_u, \Sigma_{uu})$ with 
\begin{gather}
\boldsymbol{\mu}_u = \Sigma_{um} (\Lambda_{mm}^{-1} +\Sigma_m^{-1})^{-1}\mathbf{y}
\quad \text{and}
\notag
\\
\Sigma_{uu} = K_{uu} \Delta^{-1} K_{uu}
\label{eq:fitc-regression}
\end{gather}
as mean and covariance. The matrix $\Lambda_{mm}$ is the diagonal of $K_{mm} - K_{mu} K_{uu}^{-1}K_{um}$ times identity and $\Delta = K_{uu} + (\Lambda_mm +  \Sigma_m)^{-1}  K_{mu}$.

Because of the factorized prior, the FITC approximation allows for online regression in the case where a new measurement $(\mathbf{x}_+, y_+)$ arrives. For this, the posterior distribution over $\mathbf{f}_u$ is updated from the distribution $\mathcal{N}(\boldsymbol{\mu}_u, \Sigma_{uu})$ to the distribution $\mathcal{N}(\boldsymbol{\mu}_u^+, \Sigma_{uu}^+)$ with mean and covariance
\begin{gather}
\boldsymbol{\mu}_u^+ 
=
\boldsymbol{\mu}_u + \Sigma_{u+} (\Sigma_{++} + \sigma^2)^{-1} (y_+ - \mu_+)
\quad
\text{and}
\notag
\\
\Sigma_{uu}^+
=
\Sigma_{uu} - \Sigma_{u+} (\Sigma_{++} + \sigma^2)^{-1} \Sigma_{+u}
.
\label{eq:fitc-update}
\end{gather}
The prior distribution of the new measurement enters with mean $\mu_+ = K_{+u} K_{uu}^{-1} \boldsymbol{\mu}_u$ and variance $\Sigma_{++} = K_{++} - K_{+u} K_{uu}^{-1} (K_{uu} - \Sigma_{uu}) K_{uu}^{-1} K_{u+}$, and $\Sigma_{u+} = \Sigma_{uu} K_{uu}^{-1} K_{u+}$ is the cross-covariance.

Intuitively, the FITC approximation relies on PIs in areas where measurements are located to capture the observed data. For online regression it is therefore useful to insert a new PI $\mathbf{x}_{u_+}$ to support new measurements. In this operation, the model changes from the old posterior distribution over $\mathbf{f}_u$ to the new posterior distribution over both $\mathbf{f}_u$ and $f(\mathbf{x}_{u_+})$,
\begin{equation}
\begin{pmatrix}
\mathbf{f}_u\\
f(\mathbf{x_+})
\end{pmatrix}
\sim
\mathcal{N}
\left(
\begin{pmatrix}
\boldsymbol{\mu}_u
\\
\mu_{u_+}
\end{pmatrix}
,
\begin{pmatrix}
\Sigma_{uu} & \Sigma_{uu_+}
\\
\Sigma_{u_+ u} & \Sigma_{u_+ u_+}
\end{pmatrix}
\right)
\label{eq:fitc-insert}
\end{equation}
where the mean for the new pseudo-input is $\mu_{u_+} = K_{u_+ u} K_{uu}^{-1} \boldsymbol{\mu}_u$, the variance is $\Sigma_{u_+ u_+} = K_{u_+ u_+} - K_{u_+ u} K_{uu}^{-1} (K_{uu} -  \Sigma_{uu}) K_{uu}^{-1} K_{uu_+}$, and the cross-covariance is $\Sigma_{u u_+} = \Sigma_{uu} K_{uu}^{-1} K_{uu_+}$.

If a PI $\mathbf{x} \in X_u$ should be removed, for example because it is deemed redundant, the model changes from the old posterior distribution over $\mathbf{f}_u = f(X_u)$ to the new posterior distribution over $\hat {\mathbf{f}}_u = f(\hat X_u \setminus \{ \mathbf{x} \})$. For the new posterior distribution $\mathcal{N}( \hat {\boldsymbol{\mu}}_u, \hat \Sigma_{uu})$ we only need to drop the particular dimensions, e.g.\ with the appropriate matrix multiplications
\begin{equation}
\hat {\boldsymbol{\mu}}_u = A \boldsymbol{\mu}_u
\quad
\text{and}
\quad
\hat \Sigma_{uu} =  A \Sigma_{uu} A^\intercal
\label{eq:fitc-delete}
\end{equation}
where $A$ is the identity matrix with one row removed. 

In summary, the FITC approximation provides us with a flexible regression model that allows efficient \emph{prediction} and \emph{batch initialization} with Eqs.~\eqref{eq:fitc-prediction} and \eqref{eq:fitc-regression}. The approximate regression model can be \emph{updated} with new measurements using Eq.~\eqref{eq:fitc-update} and the set of PIs can be adjusted by \emph{inserting} and \emph{removing} PIs with Eqs.~\eqref{eq:fitc-insert} and \eqref{eq:fitc-delete}. However, it is still possible that the number of PIs grows too large to be efficient and it is unclear when to insert and remove PIs. In the next section, we use the basic operations from this section and address the remaining problems by defining greedy algorithms for adjusting the set of PIs and subdividing models with too many PIs. 

\section{
\uppercase{Online Map Learning with an Ensemble of Sparse GP Experts}
}
\label{sec:method}

When learning a map of an environment with range sensors (e.g.\ dense laser scanning), observation data arrive in a sequence of measurement sets $(X_m^{(k)}, \mathbf{y}^{(k)})_{k \in \mathbb{N}}$, which contain several hundred measurements each. Already after a short interval, this results in a large and redundant amount of data and it is unrealistic to expect a single FITC GP model (see Sec. \ref{sec:FITC-gp}) to result in an efficient and accurate regression of the environment's SDF.
For this reason, we use data-driven spatial partitioning of the input space, $\mathbb{R}^2$, and create an ensemble of local experts $\{\mathbf{f}_u^{(i)}\}_i$. Each local expert $\mathbf{f}_u^{(i)} \sim \mathcal{N}(\boldsymbol{\mu}_u^{(i)}, \Sigma_{uu}^{(i)})$ has an exclusive area of responsibility $R^{(i)} \subset \mathbb{R}^2$ and its set of PIs $X_u^{(i)} = \dot{X}_u^{(i)} \dot{\cup} \ddot{X}_u^{(i)}$ consists of primary PIs $\dot{X}_u^{(i)}$ in its own area of responsibility and secondary PIs $\ddot{X}_u^{(i)} \subset \bigcup_{j \neq i} \dot{X}_u^{(j)}$ in other experts' areas of responsibility. The secondary PIs make the local experts overlap which results in more continuous predictions. To support new measurements and to ensure efficient prediction, we incrementally adapt the number of experts and their areas of responsibility.

We define the areas of responsibility based on the Voronoi diagram \citep{aurenhammer1991voronoi} of all primary PIs $\bigcup_i \dot{X}_u^{(i)}$ and set the area of responsibility of a local expert $R^{(i)}$ as the union of its primary PIs' Voronoi cells. With this spatial partitioning of the input space, we define two ways of predicting the map for a set of trail locations $X_*$. In the \emph{individual prediction} we use the posterior mean $\boldsymbol{\mu}_*^{(i)}$ from the local expert which is responsible for the respective input location,
\begin{equation}
[\boldsymbol{\mu}_*]_j =  [\boldsymbol{\mu}_*^{(i)}]_j \quad \text{where} \quad [X_*]_j \in R^{(i)}
\label{eq:individual-prediction}
\end{equation}
and in the \emph{mixture prediction} we weight the expert predictions
\begin{equation}
[\boldsymbol{\mu}_*]_j =  \sum_i \alpha_{ij} [\boldsymbol{\mu}_*^{(i)}]_j 
\label{eq:mixture-prediction}
\end{equation}
where the weights $\alpha_ij$ sum up to one for each trail location and are based on the distance between the trail input $[X_*]_j$ and the expert's primary PIs.

\subsection{Algorithm Overview}
\label{sec:overview}

When a new measurement set $(X_m^{(k)}, \mathbf{y}^{(k)})$ arrives, our learning algorithm executes the following steps for each of the local experts:
(1) Greedily extending the set of primary PIs $ \dot{X}_u^{(i)}$ where prediction error is large based on $X_m^{(k)}$ and $\mathbf{y}^{(k)}$.
(2) Updating every local expert with the new measurements.
(3) Greedily reducing the set of primary PIs $\dot{X}_u^{(i)}$ using error based on old model prediction and new measurements.
(4) Subdividing the local expert if it has too many PIs $\dot{X}_u^{(i)}$ for tractability.
(5) Harmonizing with neighbor experts by adapting the set of secondary PIs $\dot{X}_u^{(i)}$ to achieve continuous predictions. 
These steps are detailed in the sections below.

\subsection{Expert Extension}

While there exists research on choosing PIs for sparse GPs \citep{schreiter2016efficient, seeger2003fast, herbrich2003fast, smola2001sparse, csato2002sparse}, the majority of these works consider batch learning where the whole set of measurement data is available for reference, or they use sparse GP models where PI insertion changes the posterior distribution \citep{seeger2003fast, smola2001sparse, schreiter2016efficient, herbrich2003fast}. In our setting, it is however not obvious which PI will be beneficial in the long run and for simplicity we draw the candidates from the new measurement data $(X_m^{(k)}, \mathbf{y}^{(k)})$. However, we exclude all candidates outside the expert's area of responsibility and those that are closer than a minimal distance to the expert's primary PIs. 

Motivated by the maximum error criteria of \citet{schreiter2015fast}, we compare the experts posterior mean prediction for the remaining candidates $\boldsymbol{\mu}_m^{(i)}$ to their observed measurement outputs $\mathbf{y}^{(k)}$. We select the candidate $[X_m^{(k)}]_j$ with the largest error $\Vert [\mathbf{y}^{(k)}]_j - [\boldsymbol{\mu}_m^{(i)}]_j \Vert_1$ and add it to the set of primary PIs of the local expert using Eq. \eqref{eq:fitc-insert}. Before we repeat the process for the remaining candidates, we update the local expert with Eq. \eqref{eq:fitc-update} using the measurement pair $([X_m^{(k)}]_j, [\mathbf{y}^{(k)}]_j)$. In this way, new PIs are only added in areas where the current model has large error. This process is repeated until the errors fall below the threshold $t_\mathrm{add}$.

\vspace*{-1mm}
\subsection{Expert Update}
\vspace*{-1mm}

Using the new measurement set $(X_m^{(k)}, \mathbf{y}^{(k)})$, we update each local expert with Eq. \eqref{eq:fitc-update}. Measurements that are too far outside of the expert's area of responsibility or that have been used as new PIs are skipped.

\vspace*{-1mm}
\subsection{Expert Contraction}
\vspace*{-1mm}

In most cases, a local expert's predictions improve with incremental updates and PIs that were previously inserted because of their error can be removed without deteriorating the expert's SDF prediction too much. In order to maintain small sets of PIs for efficient learning and prediction, we therefore try to remove primary PIs from experts that have been extended and updated in the last step. Again, there exists research on deleting PIs from sparse GPs \citep{schreiter2015fast, schreiter2016efficient, csato2002sparse} which either does not apply to the streaming data setting or are based on a different sparse model. 

Our goal is to trade-off prediction error and model complexity when removing primary PIs. This can be achieved by comparing the posterior mean prediction for the original primary PIs $\dot{X}_u^{(i)}$ before and after the removal.
However, this evaluations only captures relative deterioration of the expert with respect to its own previous prediction. Repeatedly removing PIs up to an error threshold can therefore lead to complete deterioration. To avoid this, we additionally consider the error of posterior mean predictions on measurements from the most recent measurement set $(X_m^{(k)}, \mathbf{y}^{(k)})$ (restricted to the experts region of responsibility). We now remove the primary PI which causes the least error
\begin{equation}
\frac{1}{|X_*| + |X_{**}|}
\left\Vert  
\begin{pmatrix} 
\boldsymbol{\mu}_*^{(i)}
\\
\mathbf{y}
\end{pmatrix}
-
\begin{pmatrix} 
\hat{\boldsymbol{\mu}}_*^{(i)}
\\
\hat{\boldsymbol{\mu}}_{**}^{(i)}
\end{pmatrix}
\right\Vert_1
\end{equation}
until the error reaches the threshold $t_\mathrm{del}$ or the number of remaining primary PIs is lower than $N_\mathrm{min}$. In the equation above, $X_* = \dot{X}_u^{(i)}$, $X_{**} = X_m$ and $\hat{\boldsymbol{\mu}}_{*}^{(i)}$ and $\hat{\boldsymbol{\mu}}_{**}^{(i)}$ are the expert's posterior mean prediction for $X_*$ and $X_{**}$ when the PI is removed using Eq. \eqref{eq:fitc-delete}. This is a greedy solution to deciding the order of removed PIs, for an optimal solution, a combinatorial optimization problem which considers all possible sequences of removed PIs has to be solved.
 
\vspace*{-1mm}
\subsection{Expert Subdivision}
\vspace*{-1mm}

When a local expert covers a large area of the map or there is a lot of geometric texture in its area of responsibility, it is not possible to remove PIs without deteriorating the prediction sharply. To still maintain small numbers of PIs for efficient learning and prediction, we therefore subdivide local experts which have more than $N_\mathrm{max}$ primary PIs. For this, we cluster \citep{ward1963hierarchical} the set of primary PIs $\dot{X}_u^{(i)}$ such that no cluster is larger than $N_\mathrm{new} \leq N_\mathrm{max}$. For each cluster, we create a new local expert from the cluster's primary PIs. For the new expert's mean and covariance, we select corresponding dimensions from the old expert's $\boldsymbol{\mu}_u^{(i)}$ and $\Sigma_{uu}^{(i)}$. In this process, the secondary PIs are ignored and the new experts' predictions are potentially no longer continuous with the other experts. 

\subsection{Expert Harmonization}

When using spatial partitioning as a method of scaling GP regression to large data, it is common to ensure overlap between the local experts' measurement sets to achieve continuity in posterior mean prediction \citep{kim2013continuous}. However, in our streaming data setting, we do not have a global measurement set that we can distribute between the local experts in an overlapping manner. Instead, we ensure that the local experts' PIs $X_u^{(i)} = \dot{X}_u^{(i)} \cup \ddot{X}_u^{(i)}$ are overlapping by including primary PIs of neighbor experts as secondary PIs $\ddot{X}_u^{(i)}$. 

Since primary PI sets can change over time as described above, we need to adjust secondary PIs when the discrepancy results in discontinuous posterior predictions at the boundaries of areas of responsibility. In this case, we harmonize the local expert with its neighbors. For this, we first remove all of its secondary PIs using Eq. \eqref{eq:fitc-delete}. Next, we subsample the neighbors primary PIs to form the experts new secondary PI set. After that, we update the local expert using the new PIs and the originating neighbor's posterior mean predictions for these inputs as measurements with Eq.\ \eqref{eq:fitc-update}. This step ensures that the local expert makes similar posterior mean predictions in the neighbors' areas of responsibility as the neighbors themselves, which should result in similar predictions at the boundaries of their areas of responsibility as well. 

In order to determine when a local expert needs to be harmonized with its neighbors, we sample trail locations from the boundaries to its neighbors' areas of responsibility $X_{*j_1}, X_{*j_2}, \dots, X_{*j_K}$, which can be determined from the Voronoi diagram of all primary PIs. If the average error 
\begin{equation}
\frac{1}{\sum_{n=1}^K |X_{*j_n}|} \sum_{n=1}^K \left\Vert \boldsymbol{\mu}_{*j_n}^{(i)} - \boldsymbol{\mu}_{*j_n}^{(j_n)} \right\Vert_1
\end{equation}
between the posterior mean predictions of the local expert on a boundary $\boldsymbol{\mu}_{*j_n}^{(i)}$ and the neighbor's prediction  of this boundary $\boldsymbol{\mu}_{*j_n}^{(j_n)}$ is larger than the threshold $t_\mathrm{del}$, the local expert is harmonized.

\section{\uppercase{Experimental Results}}
\label{sec:result}

We evaluate our algorithm for learning an implicit surface model from range data in two different environments for which we have realistic synthetic or real laser range data. In all cases, we initialize the ensemble with one local expert using 50 measurements from the first range scan with Eq. \eqref{eq:fitc-regression}. If nothing else is stated, the noise variance is $\sigma^2=0.01$, the PI removal threshold is $t_\mathrm{del} = 0.01$m, the PI insertion error threshold is $t_\mathrm{add} = 0.02$m, and the minimal distance between PIs is $0.1$m. The parameters for expert size are $N_\mathrm{min} = 10$, $N_\mathrm{new} = 50$, and $N_\mathrm{max} = 100$ and every 100th range scan is used (while references to scans are still done with their original sequence number). The range parameter of the thin spline kernel \cite{williams2007gaussian} is set to $R=20$.

\begin{figure}[]
\centering
\hspace*{\fill}
\includegraphics[height=0.3\linewidth]{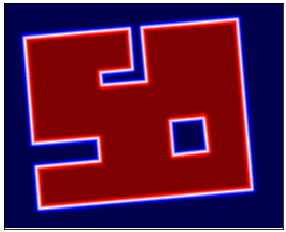}
\hfill
\includegraphics[height=0.3\linewidth]{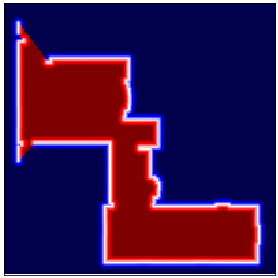}
\hspace*{\fill}

\caption{Ground truth SDF for \emph{Turtlebot} (left) and \emph{Basement-*} (right) data sets with size $26 \times 21$m and $20 \times 20$m, respectively.}
\label{fig:gt}
\end{figure}

We predict the map with a grid resolution of $0.1$m with both individual prediction (Eq. \eqref{eq:individual-prediction}) and mixture prediction (Eq. \eqref{eq:mixture-prediction}), but error measures are only evaluated for the individual prediction. We use the same error measures as reported by \citet{lee2019online}: The root-mean-square deviation (RMSD) is based on the values of the ground truth Euclidean SDF (with the same grid resolution) for map cells with predicted values between $-0.02$m and $+0.02$m. It indicates how accurately the surface is predicted on average. The Hausdorff distance ($d_H$) is computed between the center coordinates of the same map cells as in RMSD and the cells touched by the ground truth surface. It indicates an adversarial case of error and is large when a part of the surface is missed or there are large deviations. For qualitative evaluation we plot the predicted map with colors ranging from blue (negative distance) to red (positive distance) with values truncated between $-0.5$m and $+0.5$m. White color indicates the surface, as seen in Fig. \ref{fig:gt}.

\begin{figure}[]
\centering
\hfill
\includegraphics[width=0.4\linewidth]{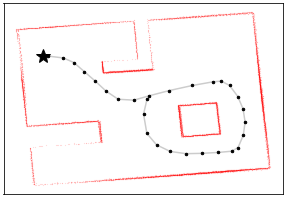}
\hfill
\setlength{\fboxsep}{0pt}
\fbox{\includegraphics[width=0.4\linewidth, trim={3mm 3mm 3mm 3mm},clip]{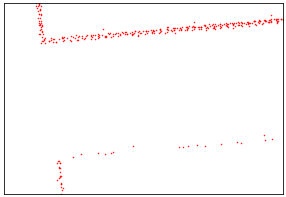}}
\hfill
\caption{\emph{Turtlebot} data set with size $26 \times 21$m (left) and a $5 \times 3.5$m section that shows details of the noise (right). The path is marked with dots and the starting position with a star.}
\label{fig:data-turtlebot}
\vspace{2mm}
\hfill
\begin{minipage}[c]{.4\linewidth}
\includegraphics[width=1\linewidth]{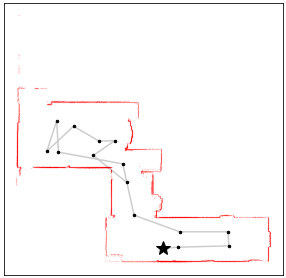}
\end{minipage}%
\hfill
\begin{minipage}[c]{0.48\linewidth}
\setlength{\fboxsep}{0pt}%
\fbox{\adjustbox{trim={.5\width} {.5\height} {.12\width} {.14\height},clip}{\includegraphics[width=1\linewidth]{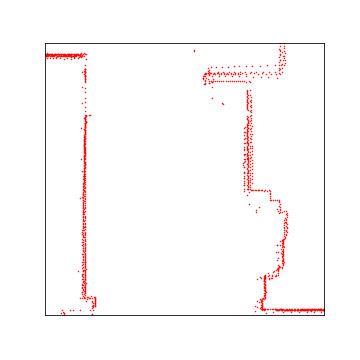}}}
\hspace{2mm}
\fbox{\adjustbox{trim={.5\width} {.5\height} {.12\width} {.14\height},clip}{\includegraphics[width=1\linewidth]{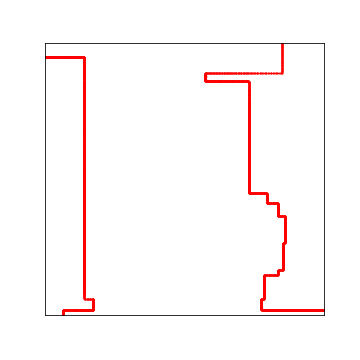}}}
\\[2mm]
\fbox{\adjustbox{trim={.5\width} {.5\height} {.12\width} {.14\height},clip}{\includegraphics[width=1\linewidth]{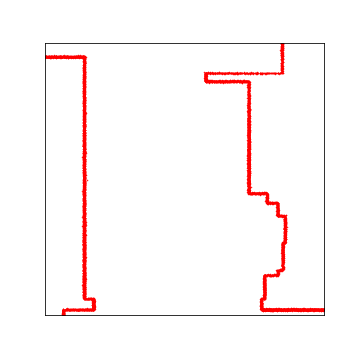}}}
\hspace{2mm}
\fbox{\adjustbox{trim={.5\width} {.5\height} {.12\width} {.14\height},clip}{\includegraphics[width=1\linewidth]{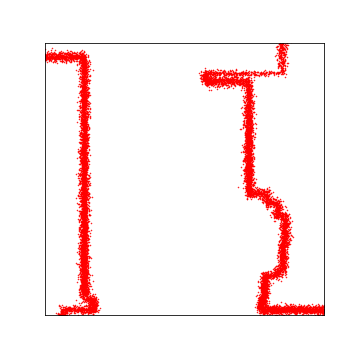}}}
\end{minipage} 
\hfill
\caption{%
\emph{Basement-real} data set with size $20 \times 20$m (left). The path is marked with dots and the starting position with a star.
\emph{Basement-*} data sets as $2.5 \times 2.5$m sections that shows details of the noise. \emph{Basement-real} (top left), \emph{Basement-GT} (top right), \emph{Basement-N} (bottom left), and \emph{Basement-NN} (bottom right).
}
\label{fig:data-basement}
\end{figure}

The first data set (\emph{Turtlebot}, 2800 scans) from \citet{lee2019online} simulates a moving Turtlebot and Hokuyo range sensor with standard deviation $0.01$m. The angular range is $-135$ degrees to $+135$ degrees and the resolution is $1$ degree.
The second data set (\emph{Basement-real}, 1500 scans) from \citet{valencia2014localization} consists of real data from a SICK S300 sensor with angular range between $-135$ degrees to $+135$ degrees and $0.5$ degree resolution. For this data set, we also have a simulated version with no sensor noise (\emph{Basement-GT, 7600 scans}), with standard deviation $0.01$m (\emph{Basement-N}, 6600 scans), and with standard deviation $0.05$m (\emph{Basement-NN}, 7000 scans). The data are described more clearly in Fig.~\ref{fig:data-turtlebot} \& \ref{fig:data-basement}.

\subsection{Baseline Comparison}

\begin{figure}[h]
\centering
\hspace*{\fill}
\includegraphics[height=0.3\linewidth]{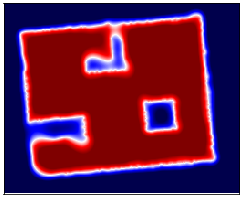}
\hfill
\includegraphics[height=0.3\linewidth]{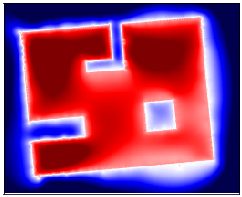}
\hspace*{\fill}
\\

\hspace*{\fill}
\includegraphics[height=0.3\linewidth]{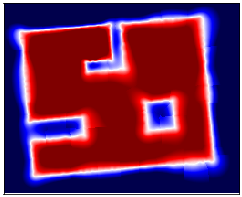}
\hfill
\includegraphics[height=0.3\linewidth]{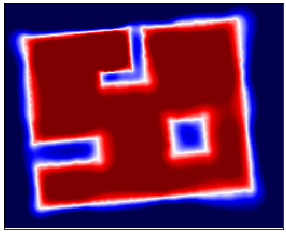}
\hspace*{\fill}
\caption{Map prediction with exact GP regression with 1000 (left) and 11000 (right) measurements (top) and individual (left) and weighted map prediction (right) with our method (bottom).}
\label{fig:individual-vs-weighted}
\end{figure}

For comparison, we compute (exact) GP regressions with the same kernel function and noise variance by subsampling 1000 to 11000 measurements from the \emph{Turtlebot} data set (15120 measurements) 10 times. 
We observe that RMSD varies between $0.04$ and $0.175$ and $d_H$ varies between $0.22$ and $1.09$ in total. 
For comparison, we run our algorithm also 10 times resulting in ensembles with $3672$ to 4068 pseudo-inputs (1138 to 1306 are primary) and RMSD between $0.054$ and $0.069$ and $d_h$ between $0.3$ and $0.59$. Compared to that, the exact regressions with 4000 measurements show RMSD between $0.049$ and $0.110$ and $d_H$ between $0.31$ and $0.59$.
For the same data set \citet{lee2019online} report RMSD of about $0.07$ and $d_H$ of about $0.22$.
Fig.~\ref{fig:individual-vs-weighted} shows renderings of the obtained map predictions for comparison. In summary, our approximate ensemble models shows similar amounts of error as exact GP regressions with smaller amount of variance over different runs and similar results to previous work.

\subsection{Sensitivity to Parameters}

We evaluate our algorithm with respect to the parameters that determine the number of PIs of local experts and set $N_\mathrm{min}$ to 10, 20, and 50, $N_\mathrm{new}$ to 50 and 100, and $N_\mathrm{max}$ to 50 and 100. 
In total, RMSD varies between $0.053$ and $0.079$ and $d_h$ varies between $0.29$ and $0.59$. The best performance is achieved for $N_\mathrm{min} = 10$ or $20$ with $N_\mathrm{new} = 50$ and $N_\mathrm{max} = 100$. Smaller $N_\mathrm{max}$ lead to larger numbers of (primary) PIs ranging form 3410 to 4680 (1125 to 1590, primary). Numerical results are listed in Fig~\ref{fig:model-size}.
This shows that our algorithm can produce similar results over a range of parameters while it is best to have $N_\mathrm{min} < N_\mathrm{new} < N_\mathrm{max}$ such that smaller experts are created by expert division and expert contraction. 

\begin{figure}[h]
\begin{center}
\tiny
\begin{tabular}{S[table-format=2.0]S[table-format=3.0]S[table-format=3.0]cS[table-format=1.2]cc} 
\toprule
$N_\mathrm{min}$ 
& 
$N_\mathrm{new}$
&
$N_\mathrm{new}$
& 
\multicolumn{1}{c}{RMSD}
&
\multicolumn{1}{c}{$d_h$}
&
\multicolumn{1}{c}{\#PI (all)}
&
\multicolumn{1}{c}{\#PI (primary)}
\\
\midrule
10 & 50 & 50 & 0.054 &0.40 & 4374 & 1384\\
10 & 50 & 100 & 0.054 & 0.29 & 3658 & 1154\\
10 & 100 & 100 & 0.079 & 0.59 & 3504 & 1125\\
20 & 50 & 100 & 0.052 & 0.31 & 4406 & 1442\\
20 & 50 & 50 & 0.056 & 0.30 & 4444 & 1458\\
20 & 100 & 100 & 0.058 & 0.50 & 3518 & 1058\\
50 & 50 & 50 & 0.053 & 0.40 & 4680 & 1590\\
50 & 50 & 100 & 0.050 & 0.30 & 4584 & 1444\\
50 & 100 & 100 &0.061 & 0.50 & 3410 & 1166\\
\bottomrule
\end{tabular}
\end{center}
\caption{Results for our algorithm when varying the parameters that determine the number of pseudo-inputs which impacts computational efficiency in learning and predicting.}
\label{fig:model-size}
\end{figure}

\begin{figure}
\centering
\hspace*{\fill}
\includegraphics[height=0.3\linewidth]{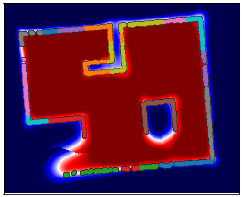}
\hfill
\includegraphics[height=0.3\linewidth]{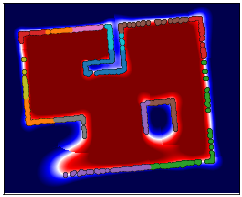}
\hspace*{\fill}
\\
\hspace*{\fill}
\includegraphics[height=0.3\linewidth]{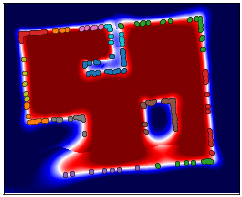}
\hfill
\includegraphics[height=0.3\linewidth]{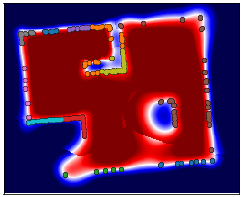}
\hspace*{\fill}
\caption{Map predictions after scan 1100 with $t_\mathrm{del} = 0.01$ (top row),  $t_\mathrm{del} = 0.02$ (bottom row)
and $t_\mathrm{add} = 0.02$ (left side), $t_\mathrm{add} = 0.05$ (right side). The primary PIs are shown as colored dots.}
\label{fig:sparsity}
\end{figure}

We also evaluate our algorithm with respect to the parameters $t_\mathrm{del}$ and $t_\mathrm{add}$ which influence the trade-off between error and model complexity. For this, we vary $t_\mathrm{del}$ between $0.001$ and $0.05$ and $t_\mathrm{add}$ between $0.005$ and $0.5$. 
As designed, we observe that larger thresholds lead to worse results with larger errors and less PIs. For visual comparison we show map predictions in Fig.~\ref{fig:sparsity} created with different threshold parameters. There, we can observe how our algorithm allocates resources (PIs): Straight, long walls use less PIs because of the thin spline kernel \cite{williams2007gaussian}, while PIs are used at corners and short walls to encode local surface curvature. The number of PIs is substantially smaller than the number of observed measurements which means that the model is a compact representation of the data. 
Numerical details about errors and number of PIs are listed in Fig.~\ref{fig:model-thresholds}.

\begin{figure}[h]
\centering
\tiny
\begin{tabular}{S[table-format=1.3]S[table-format=1.3]S[table-format=1.3]S[table-format=1.3]S[table-format=4.0]S[table-format=4.0]c} 
\toprule
$t_\mathrm{del}$ 
& 
$t_\mathrm{add}$
& 
\multicolumn{1}{c}{RMSD}
&
\multicolumn{1}{c}{$d_H$}
&
\multicolumn{1}{c}{\#PI (all)}
&
\multicolumn{1}{c}{\#PI (primary)}
\\
\midrule
0.001 
  & 0.005 & 0.058 & 0.5 & 4980 & 1786\\
" & 0.01 & 0.049 & 0.29 & 4984 & 1748\\
" & 0.02 & 0.051 & 0.3 & 4698 & 1664\\
" & 0.05 & 0.050 & 0.3 & 3796 & 1058\\
" & 0.1 & 0.073 & 0.4 & 2078 & 486\\
" & 0.2 & 0.14 & 0.59 & 1208 & 292\\
" & 0.5 & 0.479 & 1.9 & 702 & 200\\
\hline
0.005 
  & 0.005 & 0.063 & 0.5 & 4934 & 1718\\
" & 0.01 & 0.048 & 0.31 & 4918 & 1720\\
" & 0.02 & 0.05 & 0.31 & 4746 & 1674\\
" & 0.05 & 0.045 & 0.29 & 3644 & 1088\\
" & 0.1 & 0.077 & 0.59 & 1766 & 448\\
" & 0.2 & 0.12 & 0.59 & 1282 & 308\\
" & 0.5 & 0.51 & 1.9 & 686 & 190\\
\hline
0.01 
  & 0.005 & 0.057 & 0.5 & 3638 & 1146\\
" & 0.01 & 0.65 & 0.5 & 3974 & 1366 \\
" & 0.02 & 0.057 & 0.3 & 3626 & 1158\\
" & 0.05 & 0.050 & 0.3 & 3170 & 936\\
" & 0.1 & 0.074 & 0.4 & 2174 & 494\\
" & 0.2 & 0.109 & 0.5 & 1372 & 290\\
" & 0.5 & 0.384 & 1.21 & 702 & 190\\
\hline
0.02 & 0.005 & 0.182 & 1.7 & 1340 & 358\\
" & 0.01 & 0.511 & 1.90 & 968 & 198\\
" & 0.02 & 0.186 & 0.7 & 1072 & 226\\
" & 0.05 & 0.198 & 0.6 & 912 & 208\\
" & 0.1 & 0.099 & 0.70 & 1396 & 316\\
" & 0.2 & 0.132 & 0.60 & 1282 & 282\\
" & 0.5 & 0.37 & 1.4 & 684 & 190\\
\hline
0.05 & 0.005 & 0.8 & 3.1 & 22 & 22\\
\bottomrule
\end{tabular}
\caption{Results for our algorithm when varying the parameters that trade-off error and number of pseudo-inputs.}
\label{fig:model-thresholds}
\end{figure}

\subsection{Evaluation on Realistic Data Set}

\begin{figure}
\centering

\includegraphics[width=0.45\linewidth]{fig/basement_real/responsibility_1400}
\includegraphics[width=0.45\linewidth]{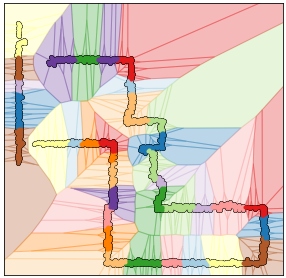}

\vspace{1mm} 

\includegraphics[width=0.45\linewidth]{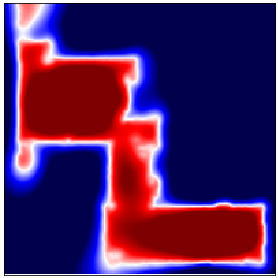}
\includegraphics[width=0.45\linewidth]{fig/basement_real/mean_combined_1400}
\caption{Final areas of responsibility of experts for \emph{Basement-real} (top left) and \emph{Basement-NN} (top right). Final map prediction for \emph{Basement-real}: weighted (bottom left) individual (bottom right).}
\label{fig:basement-responsibility}
\label{fig:basement-real-maps}

\vspace{2mm} 

\includegraphics[width=0.45\linewidth]{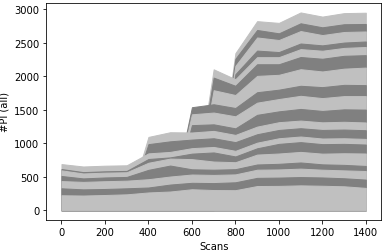}
\includegraphics[width=0.45\linewidth]{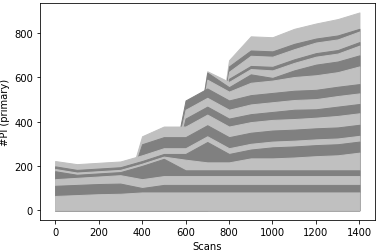}
\caption{Number (primary) pseudo-inputs over time, colors indicate different local experts.}
\label{fig:over-time}

\vspace*{2mm}

\footnotesize
\begin{tabular}{lcS[table-format=1.3]cS[table-format=4.0]cc} 
\toprule
& 
\multicolumn{1}{c}{RMSD}
&
\multicolumn{1}{c}{$d_H$}
&
\multicolumn{1}{c}{\#PI (all)}
&
\multicolumn{1}{c}{\#PI (primary)}
\\
\midrule
\emph{Basement-GT} & 0.12 & 0.09 & 2290 & 598 \\
\emph{Basement-N} &  0.06 & 0.2 & 2336 & 646 \\
\emph{Basement-NN} & 0.08 & 0.09 & 4558 & 1728 \\
\emph{Basement-real} & 0.06 & 0.09 & 2950 & 894 \\
\bottomrule
\end{tabular}
\caption{Error measures for the \emph{Basement-*} data sets based on $0.1$m grid cells in the center part of the map and one run of our algorithm.}
\label{fig:real-numbers}
\vspace{-0.6cm}
\end{figure}

We evaluate our algorithm using data from a real-world environment for which we also have simulation data with different amounts of noise. Considering real-world data allows us to judge our algorithm's practical applicability and the simulation allows us to analyze sensitivity to noise in a realistic setting. Results are shown in Fig. \ref{fig:basement-responsibility} to \ref{fig:real-numbers}.

We can see that the number of primary PIs starts at around 200 and then increases in several steps to about 900, while in between, the number also decreases sometimes, indicating expert contraction. From scan 800 on, the number of experts does not change and only primary PIs are inserted. This is the phase when the robot travels back to the starting point and gaps are filled in in unobserved locations. Local experts usually encode single geometric features such as a wall, a corner, or clutter.
From error results, we can see that increased noise leads to larger errors and larger numbers of PIs. This is because the algorithm mistakes the noise for surface details and spends more PIs to encode the structure. In this case, the  features encoded by single experts are smaller.
Finally, we also observe that \emph{Basement-real} is 300s long, and our algorithm runs for 142.74s which means that it is potentially suitable for real-time application.

\begin{figure}[]
\centering
\includegraphics[height=0.18\linewidth]{fig/basement_real/mean_combined_0}
\includegraphics[height=0.18\linewidth]{fig/basement_real/mean_combined_100}
\includegraphics[height=0.18\linewidth]{fig/basement_real/mean_combined_200}
\includegraphics[height=0.18\linewidth]{fig/basement_real/mean_combined_300}
\includegraphics[height=0.18\linewidth]{fig/basement_real/mean_combined_400}
\includegraphics[height=0.18\linewidth]{fig/basement_real/mean_combined_500}
\includegraphics[height=0.18\linewidth]{fig/basement_real/mean_combined_600}
\includegraphics[height=0.18\linewidth]{fig/basement_real/mean_combined_700}
\includegraphics[height=0.18\linewidth]{fig/basement_real/mean_combined_800}
\includegraphics[height=0.18\linewidth]{fig/basement_real/mean_combined_900}
\includegraphics[height=0.18\linewidth]{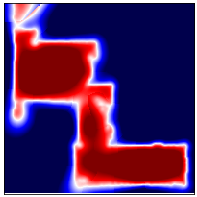}
\includegraphics[height=0.18\linewidth]{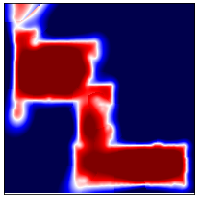}
\includegraphics[height=0.18\linewidth]{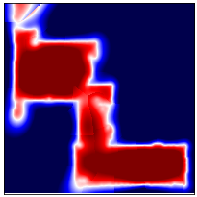}
\includegraphics[height=0.18\linewidth]{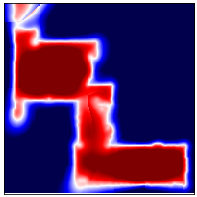}
\begin{tikzpicture}
\draw (0, 0) node[inner sep=0] {\includegraphics[height=0.18\linewidth]{fig/basement_real/mean_weighted}};
\draw (0.1, 0.61) node {\footnotesize \color{white} \textbf{Weighted}};
\end{tikzpicture}

\caption{Individual map predictions after scans 0, 100, ..., 1300 (of 1500), and the final weighted predictions for \emph{Basement-real}.}
\label{fig:map-basement-real}
\vspace{1mm}
\centering
\includegraphics[height=0.18\linewidth]{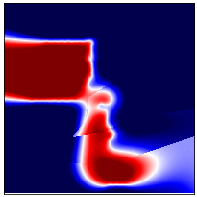}
\includegraphics[height=0.18\linewidth]{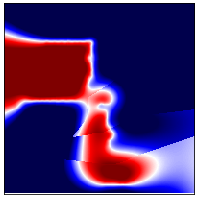}
\includegraphics[height=0.18\linewidth]{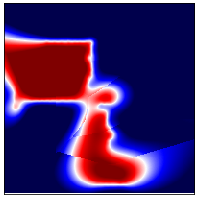}
\includegraphics[height=0.18\linewidth]{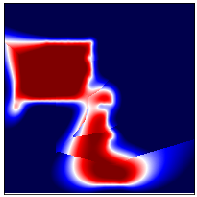}
\includegraphics[height=0.18\linewidth]{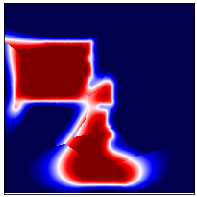}
\includegraphics[height=0.18\linewidth]{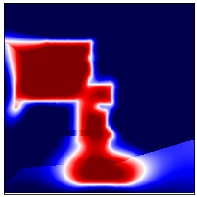}
\includegraphics[height=0.18\linewidth]{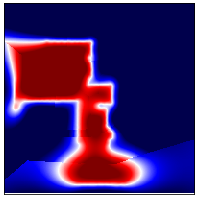}
\includegraphics[height=0.18\linewidth]{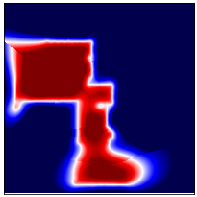}
\includegraphics[height=0.18\linewidth]{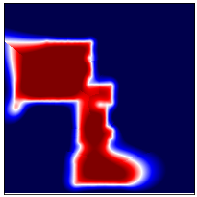}
\includegraphics[height=0.18\linewidth]{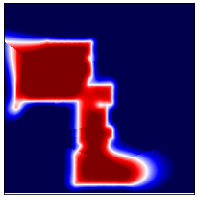}
\includegraphics[height=0.18\linewidth]{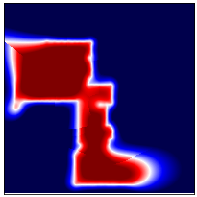}
\includegraphics[height=0.18\linewidth]{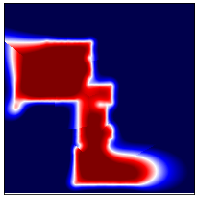}
\includegraphics[height=0.18\linewidth]{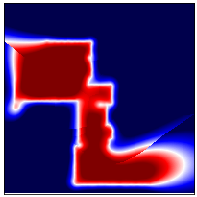}
\includegraphics[height=0.18\linewidth]{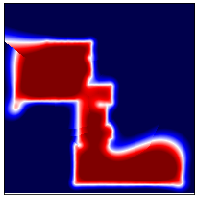}
\begin{tikzpicture}
\draw (0, 0) node[inner sep=0] {\includegraphics[height=0.18\linewidth]{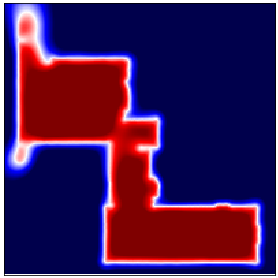}};
\draw (0.1, 0.61) node {\footnotesize \color{white} \textbf{Weighted}};
\end{tikzpicture}
\caption{Individual map predictions after scans 0, 100, ..., 1300 (of 7600), and the final weighted predictions for \emph{Basement-GT}.}
\label{fig:map-basement-gt}
\vspace{1mm}
\centering
\includegraphics[height=0.18\linewidth]{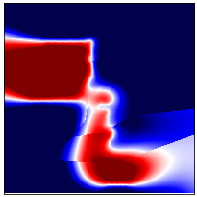}
\includegraphics[height=0.18\linewidth]{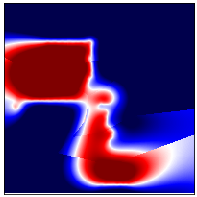}
\includegraphics[height=0.18\linewidth]{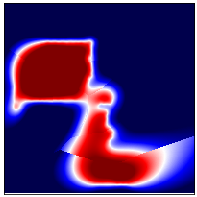}
\includegraphics[height=0.18\linewidth]{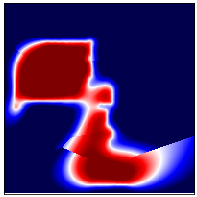}
\includegraphics[height=0.18\linewidth]{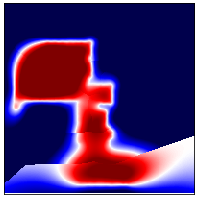}
\includegraphics[height=0.18\linewidth]{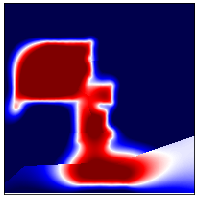}
\includegraphics[height=0.18\linewidth]{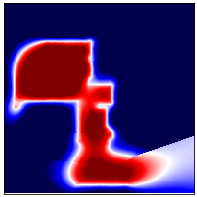}
\includegraphics[height=0.18\linewidth]{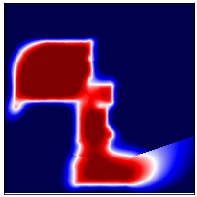}
\includegraphics[height=0.18\linewidth]{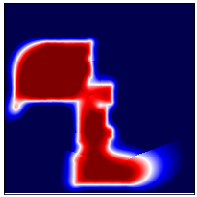}
\includegraphics[height=0.18\linewidth]{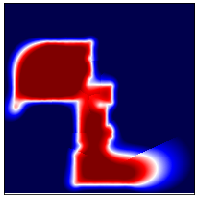}
\includegraphics[height=0.18\linewidth]{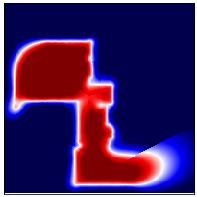}
\includegraphics[height=0.18\linewidth]{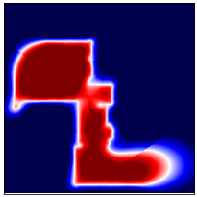}
\includegraphics[height=0.18\linewidth]{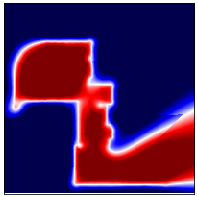}
\includegraphics[height=0.18\linewidth]{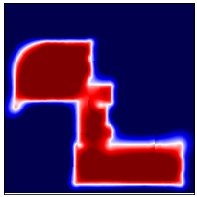}
\begin{tikzpicture}
\draw (0, 0) node[inner sep=0] {\includegraphics[height=0.18\linewidth]{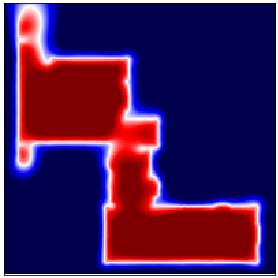}};
\draw (0.1, 0.61) node {\footnotesize \color{white} \textbf{Weighted}};
\end{tikzpicture}
\caption{Individual map predictions after scans 0, 100, ..., 1300 (of 6600), and the final weighted predictions for \emph{Basement-N}.}
\label{fig:map-basement-N}
\vspace{1mm}
\centering
\includegraphics[height=0.18\linewidth]{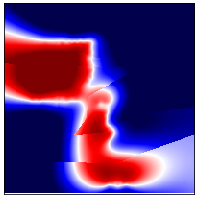}
\includegraphics[height=0.18\linewidth]{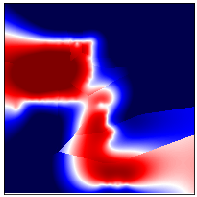}
\includegraphics[height=0.18\linewidth]{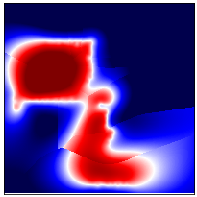}
\includegraphics[height=0.18\linewidth]{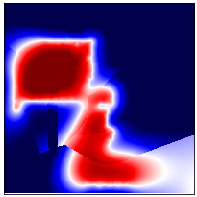}
\includegraphics[height=0.18\linewidth]{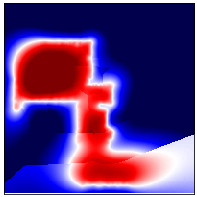}
\includegraphics[height=0.18\linewidth]{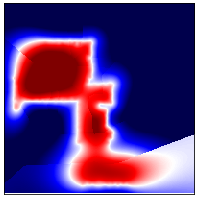}
\includegraphics[height=0.18\linewidth]{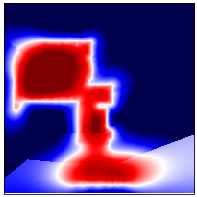}
\includegraphics[height=0.18\linewidth]{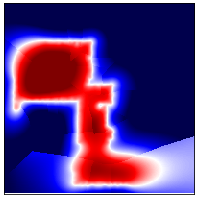}
\includegraphics[height=0.18\linewidth]{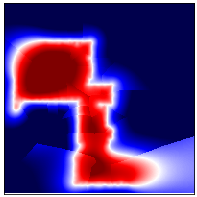}
\includegraphics[height=0.18\linewidth]{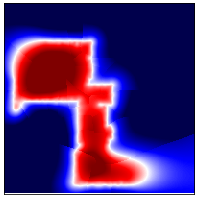}
\includegraphics[height=0.18\linewidth]{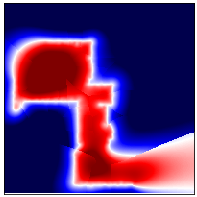}
\includegraphics[height=0.18\linewidth]{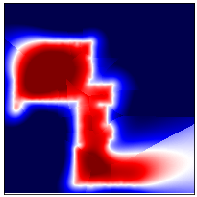}
\includegraphics[height=0.18\linewidth]{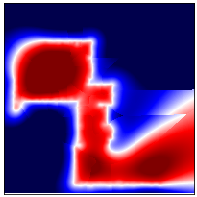}
\includegraphics[height=0.18\linewidth]{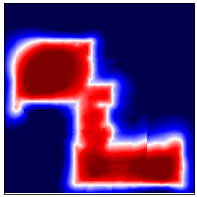}
\begin{tikzpicture}
\draw (0, 0) node[inner sep=0] {\includegraphics[height=0.18\linewidth]{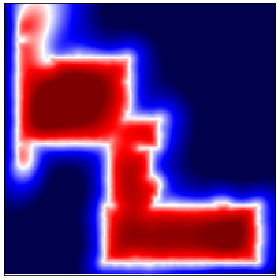}};
\draw (0.1, 0.61) node {\footnotesize \color{white} \textbf{Weighted}};
\end{tikzpicture}
\caption{Individual map predictions after scans 0, 100, ..., 1300 (of 7000), and the final weighted predictions for \emph{Basement-NN}.}
\label{fig:map-basement-NN}
\end{figure}

Fig.~\ref{fig:map-basement-real} to \ref{fig:map-basement-NN} show the individual prediction after processing scan 0 to 1300 (that means 14 scans) and the final weighted prediction. While the learning progress is different for each case because the robot moves with the same speeds, we can still observe that the general shape of the environment is learned quickly. Further, we can see that more noise leads to increasingly smoothed out features, especially in the corners, and smaller gradients of the learned distance function in regions with little data. However, in all cases, the geometric details of the environment are captured in the final prediction.

\begin{figure}[h]
\centering
\includegraphics[height=0.18\linewidth]{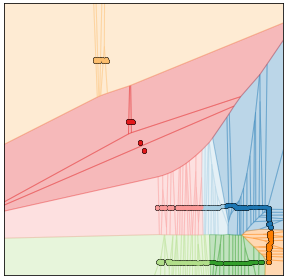}
\includegraphics[height=0.18\linewidth]{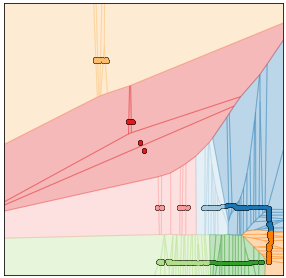}
\includegraphics[height=0.18\linewidth]{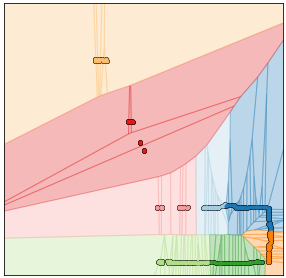}
\includegraphics[height=0.18\linewidth]{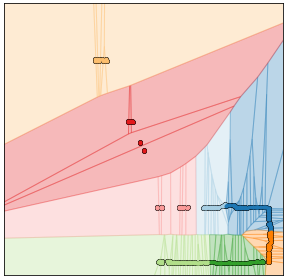}
\includegraphics[height=0.18\linewidth]{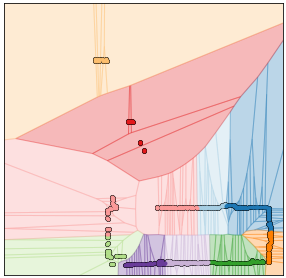}
\includegraphics[height=0.18\linewidth]{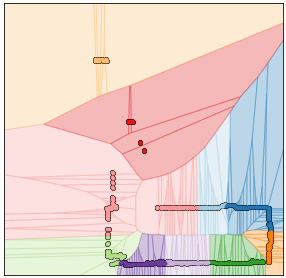}
\includegraphics[height=0.18\linewidth]{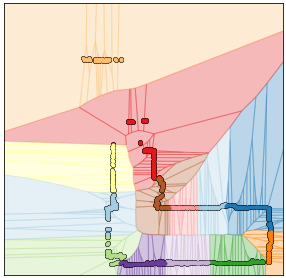}
\includegraphics[height=0.18\linewidth]{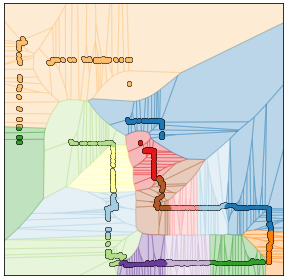}
\includegraphics[height=0.18\linewidth]{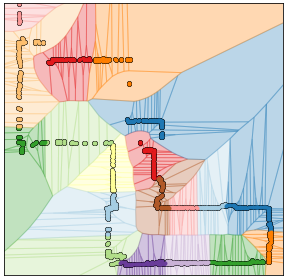}
\includegraphics[height=0.18\linewidth]{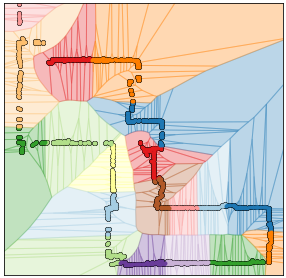}
\includegraphics[height=0.18\linewidth]{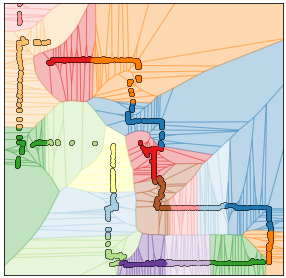}
\includegraphics[height=0.18\linewidth]{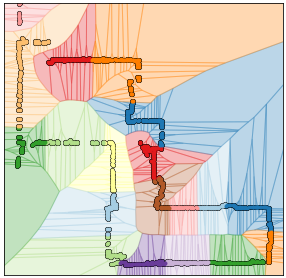}
\includegraphics[height=0.18\linewidth]{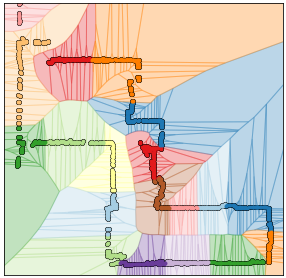}
\includegraphics[height=0.18\linewidth]{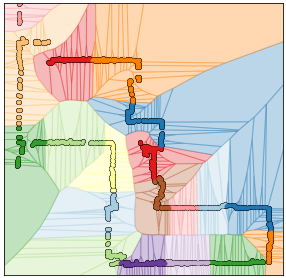}
\includegraphics[height=0.18\linewidth]{fig/basement_real/responsibility_1400}

\caption{Areas of responsibility after scans 0, 100, ..., 1400, for \emph{Basement-real} from top left to bottom right. For contrast, colors are shared between different experts.}
\label{fig:map-basement-area}
\vspace{-0.8cm}
\end{figure}

Fig.~\ref{fig:map-basement-area} shows the areas of responsibility for local experts during learning for \emph{Basement-real}. There, we can observe how local experts are subdivided and contracted when new data arrive. Furthermore, we can see that single local experts are used to capture inherent patterns of the data such as walls and corners. The local experts' PIs are grouped together and placed to capture the surface---no resources are spend to capture featureless open space.

\section{Conclusion}
\label{sec:conclusion}

In this paper, we presented an algorithm for learning an ensemble of sparse GP experts for implicit surface mapping from streaming data.
We incrementally update the expert models and adjusts model complexity---in terms of pseudo-inputs---in a greedy trade-off with prediction error. This allows learning from streaming data where an exact GP regression would be intractable. 
To achieve this, we insert and remove pseudo-inputs and subdivide expert models to maintain computational tractability. Our results from synthetic and real-world data sets show that the ensemble can predict maps with similar accuracy as exact regression and results in compact models with less pseudo-inputs in the models then measurements in the data set.
In future work, we want to extend this algorithm to 3D and dynamic environments, investigate better methods to sparsify the models, and learn hyperparameters.

\bibliographystyle{IEEEtran}
\bibliography{IEEEabrv, master}
\end{document}